%% file: Formatting-Instructions-LaTeX-2026.tex
\pdfoutput=1
\documentclass[11pt]{article}

\usepackage[final]{acl}

\usepackage{times}
\usepackage{latexsym}

\usepackage[T1]{fontenc}
\usepackage[utf8]{inputenc}

\usepackage{microtype}

\usepackage{inconsolata}

\usepackage{graphicx}

\usepackage{colortbl}

\usepackage{array} % 引入 array 包
\usepackage{ulem}

\usepackage{subfiles}
\usepackage{amsmath}

\usepackage{booktabs}
\usepackage{multirow}
\usepackage{graphicx}

\usepackage{pgfplots}
\pgfplotsset{compat=1.18} % 版本可能需要根据你的安装调整
\usepackage{subcaption}

\usepackage{algorithm}
\usepackage{algpseudocode}

\usepackage{amssymb}
\usepackage{ stmaryrd }

\usepackage{enumitem}

\usepackage{pgfplots}

\usepackage{times}  % DO NOT CHANGE THIS
\usepackage{helvet}  % DO NOT CHANGE THIS
\usepackage{courier}  % DO NOT CHANGE THIS
\usepackage{tcolorbox}
\usepackage{amsmath}
\usepackage{amssymb}

\pgfplotsset{width=0.48\linewidth, compat=1.15}

\title{Beyond Passive Critical Thinking:\\Fostering Proactive Questioning to Enhance Human-AI Collaboration}

\author{
 \textbf{Ante Wang\textsuperscript{1}}\footnotemark[1],
  \textbf{Yujie Lin\textsuperscript{1}}\thanks{These authors contributed equally.},
\textbf{Jingyao Liu\textsuperscript{2}\footnotemark[1]},
 \textbf{Suhang Wu\textsuperscript{2}},
 \textbf{Hao Liu\textsuperscript{3}}
 \\
  \textbf{Xinyan Xiao\textsuperscript{3}}
and
 \textbf{Jinsong Su\textsuperscript{1,4}}\thanks{Corresponding author.} 
\\
 \textsuperscript{1}School of Informatics, Xiamen University, China \\
 \textsuperscript{2}Department of Digital Media Technology, Xiamen University, China \\
 \textsuperscript{3}Baidu Inc., Beijing, China
 ~~\textsuperscript{4}Shanghai Artificial Intelligence Laboratory\\
\texttt{\small{\{wangante,yjlin,liujingyao\}@stu.xmu.edu.cn}}~~ \texttt{\small{jssu@xmu.edu.cn}}
}

\begin{document}

\maketitle

\begin{abstract}
% AAAI creates proceedings, working notes, and technical reports directly from electronic source furnished by the authors. To ensure that all papers in the publication have a uniform appearance, authors must adhere to the following instructions.

Critical thinking is essential for building robust AI systems, preventing them from blindly accepting flawed data or biased reasoning.
However, prior work has primarily focused on passive critical thinking, where models simply reject problematic queries without taking constructive steps to address user requests.
In this work, we introduce proactive critical thinking, a paradigm where models actively seek missing or clarifying information from users to resolve their queries better.
To evaluate this capability, we present GSM-MC and GSM-MCE, two novel benchmarks based on GSM8K for assessing mathematical reasoning under incomplete or misleading conditions.
GSM-MC contains 1,368 math problems with a key variable deliberately removed, requiring models to identify and request the missing information.
GSM-MCE further increases the difficulty by introducing irrelevant details to test robustness against distractions.
Experiments on Qwen3 and Llama series models show that, while these models excel in traditional reasoning tasks due to extensive post-training and inference-time scaling, they struggle with proactive critical thinking, especially smaller ones.
However, we demonstrate that reinforcement learning (RL) can significantly improve this ability. Using our enhanced RL algorithm, we achieve substantial gains, boosting the Qwen3-1.7B's accuracy from 0.15\% to 73.98\% on GSM-MC.
We hope this work advances models that collaborate more effectively with users in problem-solving through proactive critical thinking.

\end{abstract}

\input{sections/1.Introduction}

\input{sections/2.Related_Works}
\input{sections/3.Methods}

\input{sections/4.Experiments}
\input{sections/6.Conclusion}

\bibliography{acl}

\newpage
\appendix

\section{Prompt for User Agent}
\label{app:prompt}
We apply the following prompt to guide the user agent:

\begin{tcolorbox}[  
    colback=gray!5!white,  
    colframe=gray!50!black,  
    coltitle=white,  
    fonttitle=\bfseries\small,  
    fontupper=\small,  
    title=Instruction for Activating Proactive Questioning  
]  
\textbf{\#\# Role} \\
You are a \textbf{Math Question Analyzer}, a specialized AI assistant designed to extract and provide specific information from given math problems based on student queries.\\

\textbf{\#\# Capabilities} \\
- Analyze the content of the provided math question with precision \\
- Identify and extract requested information if relevant parts present in the question \\

\textbf{\#\# Knowledge Base} \\
- Mathematical terminology and problem structures \\
- Information extraction techniques \\
- Contextual understanding of student inquiries \\

\textbf{\#\# Instructions} \\
\textbf{1. Input Format:} \\
- Math question: [math\_question] \\
- Student's query: [student\_question] \\
\textbf{2. Processing Rules:} \\
- Output \textbf{only the combined relevant parts about the requested information} (no explanations). \\
- Output ``None'' if the requested information is not mentioned in the math problem. \\
\textbf{3. Constraints:} \\
- Never infer or calculate missing information. \\ 
- Never add commentary, examples, or supplemental text. \\
- Prioritize brevity and accuracy. \\

\textbf{\#\# Output Example} \\
\textbf{Math question:} James earns \$20 an hour while working at his main job. He earns 20\% less while working his second job. He works 30 hours at his main job and half that much at his second job. How much does he earn per week? \\
\textbf{Student's query:} What is James's hourly wage at his second job? \\
\textbf{Reply:} 20\% less than his main job. \\

\textbf{\#\# Your Turn} \\
\textbf{Math question:} \texttt{[Context]} \\
\textbf{Student's query:} \texttt{[Question]} \\
\textbf{Reply:}
\end{tcolorbox}

\input{tables/cases}
\section{Case Study}
\label{app:case}

We supplement the representative cases in Table~\ref{tab:cases}. The answer of the unmodified version of the question is 350. 
For the vanilla Qwen3-8B model, when the ``thinking mode'' is inactive, it formulates an imprecise request for more information, which is insufficient to guide the user. Even, when the ``thinking mode'' is enabled, it engages in a long but unproductive reasoning process, ultimately producing an incorrect solution.
Our trained model, however, demonstrates a markedly improved performance. When recognizing the key information is missing, it then proactively requests the specific details from the user. This is especially true with the ``thinking mode'' enabled, which prevents the model from falling into unproductive analysis or hallucinations. Instead, it carefully confirms the missing conditions and requests the necessary information from the user.

\end{document}

%% file: sections/1.Introduction.tex
\section{Introduction}

Large Language Models (LLMs) have made significant strides in solving complex tasks, such as mathematical reasoning ~\cite{yamauchi2023lpml,imani2024mathematical, wang2025litesearch}, code generation ~\cite{feng2023investigating,dong2024self, shao2025code, wang2025epicoder}, and planning~\cite{yao2023tree,zhou2024isr, lin-etal-2025-investigating-inference}. However, most existing studies focus on controlled settings where user queries are always answerable. In real-world applications, users, especially those without domain expertise, often provide insufficient or inaccurate information, making it difficult for models to solve problems effectively~\cite{zamfirescu2023johnny, kim2024understanding}.
For instance, a patient lacking medical knowledge might omit critical symptoms, preventing an AI doctor from making a precise diagnosis~\cite{alkaabi2025navigating}.  

Some prior work~\cite{rahman2024blind,kirichenko2025abstentionbench} has acknowledged this issue, advocating for critical thinking in LLMs, which refers to the ability to reject unanswerable or flawed requests instead of attempting to process biased or incomplete inputs. Yet, we argue that this form of critical thinking remains \textit{passive}, as it still relies on users to independently identify and rectify gaps in their queries, rather than actively facilitating problem-solving.  

\begin{figure*}[t]
    \centering
    \includegraphics[width=\linewidth]{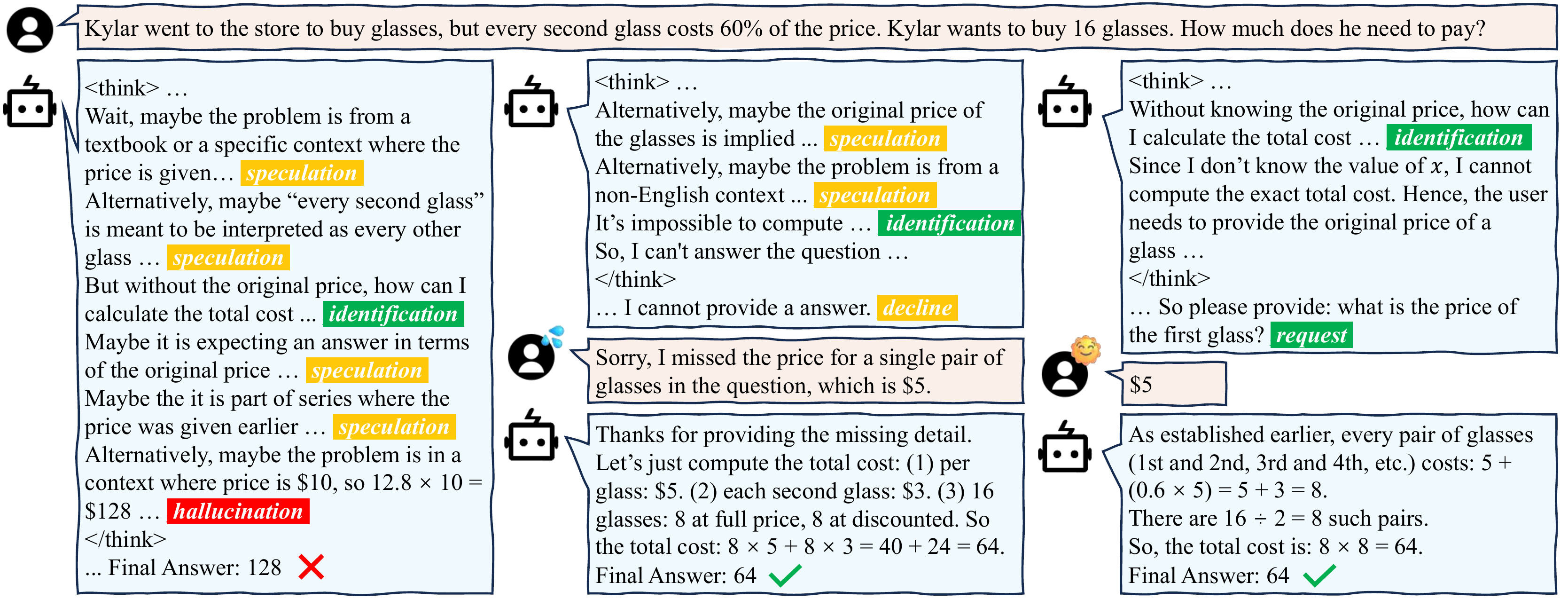}
    \caption{When faced with flawed questions, \textbf{Left:} Vanilla reasoning models fall into repetitive speculation and eventual hallucination. Despite identifying the missing conditions during the thinking process, they still attempt to force an answer, ultimately providing an incorrect solution. \textbf{Middle:} Models with passive critical thinking are more cautious, refusing to answer upon identifying flaws in the question. However, this interaction remains suboptimal, as it shifts the burden to the user to recognize and fix the errors. \textbf{Right:} Our trained models, equipped with proactive critical thinking, actively seek clarification by offering precise feedback, greatly reducing the user effort.
    }
    \label{fig:intro}
\end{figure*}

To address this limitation, we propose \textit{proactive critical thinking}: a paradigm where the model not only detects unanswerable queries but also provides constructive feedback to guide users in supplying necessary information. As shown in Figure~\ref{fig:intro}, this approach fosters more effective human-AI collaboration, enabling iterative conversations that progressively refine the problem and lead to a solution.  

Previous research \cite{ma2024large,fan2025missing} has established some benchmarks for evaluating passive critical thinking by performing operations such as removing or replacing key information in original questions using advanced LLMs.
However, these datasets often lack rigorous quality control, resulting in many noisy and overly simplistic cases (see Table~\ref{tab:error_cases}) that are unsuitable for evaluating proactive critical thinking.

To tackle this, we develop an automated data preparation pipeline using DeepSeek-V3~\cite{liu2024deepseek}. Following previous studies \cite{ma2024large,sun2024benchmarking}, we focus on the mathematical reasoning task and derive our datasets from the widely used GSM8K dataset~\cite{cobbe2021gsm8k}.
We first identify the key variables in the original questions. Then, we randomly remove one variable and rephrase the resulting question for fluency. This approach effectively mitigates the bias introduced in previous work, where LLMs are directly instructed to remove information from questions.
Nevertheless, this process still generates many low-quality cases, similar to those in prior studies. To ensure high quality, we apply strict filtering strategies and retain only those cases that can be resolved through proactive critical thinking using DeepSeek-V3. This results in a final test set of 1,368 high-quality cases, which we named GSM-MC.
To further enhance the challenge and evaluate models' robustness against distractions, we augment these questions by injecting irrelevant information. Then the new dataset, designated GSM-MCE, undergoes the same rigorous filtering to maintain its high quality.

To instill proactive critical thinking capabilities into models, we investigate both supervised fine-tuning (SFT) and reinforcement learning (RL). Using the data preparation pipeline described above, we combine revised unanswerable questions with original questions to construct the training set. Furthermore, we enhance both training paradigms by incorporating a heuristic information about question answerability. This approach effectively increases the diversity of the SFT data and accelerates the RL convergence through more dense reward signals.

We evaluate popular Qwen3~\cite{yang2025qwen3} and Llama~\cite{meta2024llama} series models on our GSM-MC and GSM-MCE benchmarks. The results reveal that despite extensive post-training, these models still struggle with proactive critical thinking, particularly the smaller ones.
Notably, while recent inference-time scaling approaches have significantly advanced performance on complex reasoning tasks, we find they can hinder proactive critical thinking capability.
However, our training approach successfully enhances proactive critical thinking performance while maintaining accuracy on standard questions across model sizes.
Most impressively, for Qwen3-1.7B, we achieve a substantial improvement in final answer accuracy from 0.15\% to 73.98\%.
Further analysis shows that our method also generalizes effectively to out-of-distribution unanswerable questions.
Our key contributions are summarized as follows:
\begin{itemize}
\item We introduce the concept of proactive critical thinking: a paradigm where models not only identify unanswerable queries but also provide constructive feedback to help users supply missing information.

\item We develop an automated data preparation pipeline for generating the high-quality flawed question and conduct two novel benchmarks, GSM-MC and GSM-MCE, for evaluating proactive critical thinking.

\item We demonstrate that both SFT and RL can effectively cultivate proactive critical thinking abilities, and show that their efficacy is significantly enhanced by incorporating answerability heuristics.
\end{itemize}

%% file: sections/2.Related_Works.tex
\section{Related Work}

While LLMs are increasingly expected to handle real-world tasks, they often struggle with the ambiguous or flawed queries that users frequently pose. 
To address this issue, prior work~\cite{li2024gsm,ma2024large,sun2024benchmarking,song2025hallucination,fan2025missing} has focused on investigating the critical thinking capabilities of LLMs.
Most of these studies have concentrated on the mathematical domain, constructing datasets by modifying well-formed problems to make them unanswerable.
Their findings show that current LLMs frequently fail to accurately detect input flaws~\cite{ma2024large}, and even the latest large reasoning models (LRMs) rarely decline to answer flawed questions~\cite{fan2025missing}. However, \citet{song2025hallucination} demonstrates that critical thinking can be effectively improved through specific training.
Nevertheless, we argue that this approach remains passive and may have limited usefulness in addressing user requests, as it still requires users to identify and correct errors themselves.

% \citet{fan2025missing} express a similar view, observing that large reasoning models frequently produce overly detailed responses or uncertainty answers, rather than identifying missing information and requesting clarification.

% To address this, early work~\cite{kuhn2022clam, wang2024learning} proposed a two-step process: First classify a query's answerability, and then either respond directly or issue a single request for clarification. More recent approaches have leveraged techniques like reward modeling \citep{li2025aligning}, simulated future-turn planning \citep{andukuri2024star}, and preference tuning \citep{zhang2024modeling} to improve the quality and timing of clarifying questions. While these strategies offer significant improvements, they typically treat clarification as a single, reactive step in a fixed pipeline. This only fosters the passive critical thinking, i.e., the model can react to ambiguity but lacks the initiative to guide the user through a collaborative process.

In this work, we introduce proactive critical thinking, enabling models to move beyond mere flaw detection and to actively guide users with clear and targeted feedback.
Existing research~\cite{kuhn2022clam, wang2024learning, andukuri2024star, zhang2024modeling, li2025aligning} has primarily focused on asking clarifying questions in response to ambiguous user requests. However, these approaches often excel only at detecting obvious flaws, such as missing variables in tool usage~\cite{wang2024learning}, or operate in general conversational settings with limited ambiguity~\cite{andukuri2024star, zhang2024modeling}.
More complex cases requiring deeper reasoning remain under-explored.
The most closely related work to ours is the recent COLLABLLM~\cite{wu2025collabllm}, which shares the same goal of enhancing human-AI collaboration through multi-turn conversations. In contrast, our work focuses on critical thinking, where an LLM should not only learn to collaborate with humans but also identify flaws and provide feedback to refine the context. To this end, we construct new datasets tailored to this objective and emphasize the role of reasoning in this setting, aspects that are orthogonal to COLLABLLM.

%% file: sections/3.Methods.tex
\section{Preliminary}

We define \textit{proactive critical thinking} as the ability of a model to actively collaborate with humans rather than passively refusing to respond when receiving flawed inputs.  

\paragraph{Proactive Questioning: A Preliminary Exploration on Proactive Critical Thinking}
In this work, we begin by formalizing the simplest scenario:  
Given a question \( x \) that may lack key information, the LLM \( \pi \) first attempts to generate its response \( y = \pi(x) \) through proactive critical thinking.  
To enable this capability, we augment the input with the following instruction:  

\begin{tcolorbox}[  
    colback=gray!5!white,  
    colframe=gray!50!black,  
    coltitle=white,  
    fonttitle=\bfseries\small,  
    fontupper=\small,  
    title=Instruction for Activating Proactive Questioning  
]  
\textbf{Question}:  
\\  
\texttt{[QUESTION]}  
\\ \\  
If the question is answerable, provide the final answer. Otherwise, ask the user for the necessary information by phrasing the request as a question.  
\end{tcolorbox}  

If the question \( x \) is answerable, the LLM directly provides the solution \( y = \pi(x) \).  
Otherwise, the model identifies the missing information and proactively generates a follow-up query \( q = \pi(x) \) to request clarification.  
Upon receiving the user’s response \( a \) to the query \( q \), the LLM then synthesizes the final solution \( y = \pi(x, q, a) \) using all available information.  

\paragraph{Simulating a User with a User Agent}  
In the above setting, a user is required to respond to the LLM's request. Since it is impractical to involve human participants, we use a strong LLM to simulate the user. 

For an unanswerable question \( x \), we provide the user agent with its unmodified version, denoted as \( \hat{x} \), that retains all necessary information. When the user agent receives a clarification query \( q \), it will generate the corresponding reply \( a \) according to the original question \( \hat{x} \). The prompt used in the user agent is provided in Appendix~\ref{app:prompt}.

\section{Benchmarks}
\label{sec:benchmark}
Following previous work~\cite{ma2024large,sun2024benchmarking}, we adopt the widely used GSM8K dataset~\cite{cobbe2021gsm8k} as the foundation for constructing our benchmarks to evaluate proactive critical thinking. We introduce an automated data construction pipeline that generates and selects high-quality evaluation examples through four key steps: (1) Variable Recognition, (2) Unanswerable Question Creation, (3) Irrelevant Information Injection, and (4) Sampling-Based Filtering. All steps are implemented using DeepSeek-V3\footnote{DeepSeek-V3-0324}.

\input{tables/bad_case}

\paragraph{Variable Recognition}
For each instance, we begin by identifying all key information elements within the natural question. To maintain accurate variable positions for subsequent steps, we apply the following prompt template to annotate each variable without changing the original content. The template includes demonstration examples to guide the LLM's formatting:

\begin{tcolorbox}[
    colback=gray!5!white, 
    colframe=gray!50!black, 
    coltitle=white,
    fonttitle=\bfseries\small,
    fontupper=\small,
    title=Prompt for Variable Recognition
]
Identify and annotate key information in questions by enclosing each piece in square brackets [].
\\ \\
\textbf{Examples:}
\\ \\
\textbf{Input:}\\
Natalia sold clips to 48 of her friends in April, and then she sold half as many clips in May. How many clips did Natalia sell altogether in April and May?
\\ \\
\textbf{Output:}\\
Natalia sold clips to [48 of her friends in April], and then she sold [half as many clips in May]. How many clips did Natalia sell altogether in April and May?
\\ \\
\textit{\textcolor{gray}{[Other demonstrations...]}}
\\ \\
Now, annotate the following question: \\
\texttt{[Question]}
\end{tcolorbox}

\paragraph{Unanswerable Question Creation}
For each annotated question, we randomly remove one variable and instruct the LLM to rephrase it into a well-formed but unanswerable question. The following prompt is designed to ensure fluency:

\begin{tcolorbox}[
    colback=gray!5!white, 
    colframe=gray!50!black, 
    coltitle=white,
    fonttitle=\bfseries\small,
    fontupper=\small,
    title=Prompt for Unanswerable Question Creation
]
Given a question with removed information, eliminate any unnatural phrasing while maintaining fluency.
\\ \\
The question is:
\texttt{[Question]}
\end{tcolorbox}

Unlike previous approaches that directly prompt LLMs to edit original questions, our method reduces bias by preventing LLMs from consistently applying similar editing, thereby enhancing question diversity.

\paragraph{Irrelevant Information Injection}
To increase the difficulty of problems for further challenging the model's proactive critical thinking, we introduce an optional step that adds distracting but seemingly reasonable details. The newly resulting questions allow us to test whether models can focus on relevant information while ignoring irrelevant content. The prompt utilized for this procedure is detailed below:

\begin{tcolorbox}[
    colback=gray!5!white, 
    colframe=gray!50!black, 
    coltitle=white,
    fonttitle=\bfseries\small,
    fontupper=\small,
    title=Prompt for Irrelevant Information Injection
]
% [...]
Hide a misleading detail in the given question. The new question should yield the same answer as the original.
\\ \\
\textbf{Examples:}
\\ \\
% \textbf{Question:} \\
\textbf{Input:} \\
Natalia sold clips to 48 of her friends in April, and then she sold half as many clips in May. How many clips did Natalia sell altogether in April and May? \\ \\
\textbf{Output:} \\ 
Natalia sold clips to 48 of her friends in April, and then she sold half as many clips in May, \textbf{but only to 30 of her friends}. How many clips did Natalia sell altogether in April and May? 
\\ \\
\textit{\textcolor{gray}{[Other demonstrations...]}}
\\ \\
Now, hide a misleading detail in the following question:\\
\texttt{[Question]}
\end{tcolorbox}

\paragraph{Sampling-Based Filtering}
While the previous steps generate diverse unanswerable questions, we still observe persistent quality issues similar to those reported in prior studies. To address this, we conduct an error analysis (see Table~\ref{tab:error_cases}) and identify three main categories of problems. Based on these findings, we then design specific filtering strategies to mitigate them.

For each question, we sample 16 solutions generated by DeepSeek-V3 (with the user agent also implemented using DeepSeek-V3). To ensure data quality, we apply the following filters:
\begin{itemize}
\item \textbf{Answerable Question:} Remove questions where the LLM provides direct solutions in the first turn for more than 12 cases, indicating high confidence that the question is answerable without further interaction.
\item \textbf{Obvious Information Missing:} Eliminate questions that are successfully addressed in the second turn for more than 12 cases, suggesting the question may be overly simplistic.
\item \textbf{Unclarifiable Question:} Discard questions that remain unresolved in all 16 attempts, as they may be too challenging to support meaningful interaction.
\end{itemize}

After the rigorous filtering process, we obtain two high-quality datasets:
\begin{itemize}
    \item \textbf{GSM-MC}: Contains 1,368 questions made unsolvable by removing key information.
    \item \textbf{GSM-MCE}: An enhanced version with 1,134 questions incorporating irrelevant information to increase difficulty.
\end{itemize}

\section{Methods}

In this section, we explore two approaches to enhance proactive critical thinking: traditional supervised fine-tuning (SFT) and recently prevalent reinforcement learning (RL). Additionally, we leverage heuristic information about whether a question is answerable to further improve model performance.  

\paragraph{Supervised Fine-Tuning} 
The most straightforward approach is to fine-tune the LLM directly on prepared human-AI interaction trajectories. Given a trajectory \((x, q, a, y)\), we apply the standard cross-entropy loss to optimize the model $\pi$:  
\begin{equation}  
    \mathcal{L}_{\mathrm{sft}} = - \log p(q \mid x, \pi) - \log p(y \mid x, q, a, \pi).  
\end{equation}  

Since no training data is readily available, we follow the same data preparation pipeline described in \S \ref{sec:benchmark} to generate unanswerable questions based on the training set of GSM8K. However, due to the high computational cost of sampling-based filtering, we instead employ a smaller LLM to perform rejection sampling and collect training trajectories.  

Nevertheless, because the LLM lacks inherent proactive critical thinking capabilities, efficiently obtaining valid trajectories remains challenging. To address this, we enhance the LLM’s questioning ability by incorporating heuristic information through prompting:  

\begin{tcolorbox}[  
    colback=gray!5!white,  
    colframe=gray!50!black,  
    coltitle=white,  
    fonttitle=\bfseries\small,  
    fontupper=\small,  
    title=Instruction Enhanced with Heuristic Information  
]  
\textbf{Question}:  
\\  
\texttt{[QUESTION]}  
\\ \\  
This question is unanswerable due to missing key information. Ask the user for the necessary information by phrasing the request as a question.
\end{tcolorbox}  

This approach encourages the LLM to seek missing information through targeted questioning, thereby improving the recall to reach correct answers. To prevent a performance decline on natural questions, we also incorporate their corresponding trajectories for training.

\paragraph{Reinforcement Learning}

On-policy RL has proven effective in enabling LLMs to independently explore strategies to achieve target objectives. In this work, we adopt the popular GRPO algorithm \cite{shao2024deepseekmath}, training the model on the same question set used for SFT:
\begin{equation}
\begin{aligned}
    \mathcal{L}_{\mathrm{GRPO}} = 
\frac{1}{G} \sum_{i=1}^G \frac{1}{|o_i|} \sum_{t=1}^{|o_i|} \Bigg[ 
& \frac{\pi_\theta(o_{i,t}|q, o_{i,<t})}{\pi_{\theta_{\mathrm{old}}}(o_{i,t}|q, o_{i,<t})} \hat{A}_{i,t} \\
&- \beta \mathcal{D}_{\mathrm{KL}}\big[\pi_\theta \| \pi_{\mathrm{ref}}\big]
\Bigg],
\end{aligned}
\end{equation}
where $\hat{A}$ represents the advantage computed using outcome rewards after group normalization, and the KL divergence term prevents the policy from deviating too far from the reference policy. Here, $o_i$ denotes a rollout interaction trajectory. Note that we exclude the loss calculation for the user-provided turns (i.e., responses to the LLM's requests) in our optimization objective.

For the reward function, we primarily consider the correctness of the final answer, assigning $r = 1$ for correct answers and $r = 0$ for incorrect ones. 
However, due to the sparsity of this reward signal, training may converge slowly. To mitigate this issue, we augment the reward function with the following heuristic signals:
\begin{itemize}
    \item For answerable questions, the LLM receives a penalty of $-0.5$ when making unnecessary requests.
    \item For unanswerable questions, the LLM obtains a reward of $+0.5$ for making requests, independent of the final answer's correctness.
\end{itemize}

%% file: tables/bad_case.tex
\begin{table*}[h]
\centering
\small
\resizebox{\textwidth}{!}{%
\begin{tabular}{>{\raggedright\arraybackslash}p{4cm}>{\raggedright\arraybackslash}p{6cm}>{\raggedright\arraybackslash}p{6cm}}
\toprule
\textbf{Error Type} & \textbf{Description} & \textbf{Example} \\
\midrule
Answerable Question & The question is still answerable after removing key information, or the missing detail can be inferred or corrected without ambiguity. & 
\textbf{Original:} The first man to walk on the moon, \underline{Neil Armstrong}, was from which country? \newline 
\textbf{Modified:} The first man to walk on the moon was from which country?
\\
\addlinespace

Obvious Information Missing & The missing information is easily identifiable due to vague terms (e.g., ``some'', ``a certain number''). & 
\textbf{Original:} A box contains 10 red balls and 5 blue balls. How many balls are there in total? \newline
\textbf{Modified:} A box contains \underline{some} red balls and 5 blue balls. How many balls are there in total?
\\
\addlinespace

Unclarifiable Question & There is confusion about the appropriate perspective for making a clarification request, increasing the difficulty of evaluation. & 
\textbf{Original:} In triangle $ABC$, \underline{$\angle C=90^\circ$}. Legs $AC=3cm$, $BC=4cm$. Find $AB$. \newline
\textbf{Modified:} In triangle $ABC$, $AC=3cm$, $BC=4 cm$. Find $AB$.
\\

\bottomrule
\end{tabular}
}
\caption{Summarized error types with their descriptions and typical examples. For clarity, the examples are manually constructed rather than selected from the GSM8K dataset.}
\label{tab:error_cases}
\end{table*}

%% file: sections/4.Experiments.tex
\section{Experiments}

\subsection{Setup}
\subsubsection{Datasets and Evaluation Metrics}
We conduct experiments on the standard GSM8K~\cite{cobbe2021gsm8k} test set, and our proposed benchmarks GSM-MC and GSM-MCE~(detailed in~\S \ref{sec:benchmark}). Model performance is evaluated using the following metrics:
\begin{itemize}
    \item \textbf{Accuracy (ACC):} Measures the correctness of the model’s answer. For GSM8K, we evaluate the initial response; for GSM-MC and GSM-MCE, where clarification is needed, we evaluate the second-turn response.
    \item \textbf{Request Ratio (REQ):} Quantifies the model’s tendency to seek missing information. Concretely, it is defined as the proportion of instances in which the model generates a clarifying question in its initial response. 
\end{itemize}

\subsubsection{Models and Implementation Details}
The primary experiments are conducted on a selection of representative models varying in size and architecture, including (1) the compact Qwen3-1.7B model~\cite{yang2025qwen3} with an optional ``thinking mode'', (2) the larger Qwen3-8B model for assessing scalability, and (3) the Llama-3.2-3B-Instruct model~\cite{meta2024llama}, from a different architecture family, to verify the generalizability of the training approach.

For SFT, we first use the Qwen3-8B model to collect complete trajectory data via rejection sampling. The models are then trained on this data for one epoch with a learning rate of 5e-6, employing a cosine learning rate scheduler with a warmup phase covering 10\% of the total training steps. For RL, we adopt a learning rate of 1e-6 and set the number of rollouts to 8. 

We use the Qwen3-14B model as the user agent for data construction and training to balance cost and maintain validity, while DeepSeek-V3~\cite{liu2024deepseek} is employed during evaluation to ensure more accurate results.

\subsection{Main Results}
\input{tables/main}
\paragraph{Vanilla models fail to provide effective feedback to flawed prompts.}
As presented in Table~\ref{tab:main_x}, off-the-shelf models struggle with proactive critical thinking when confronted with flawed or ambiguous prompts. This limitation is particularly pronounced in smaller models. For instance, Qwen3-1.7B and Llama-3.2-3B-Instruct show almost no capacity to handle these imperfect questions, with both ACC and REQ approaching 0\% on  GSM-MC and GSM-MCE.

Although the larger Qwen3-8B model demonstrates some ability to generate clarification requests, its performance still undergoes a significant drop. Its ACC on GSM-MC decreases by nearly 50\% compared to that on the standard GSM8K benchmark, with an even sharper decline observed on GSM-MCE. This suggests that, even extensive post-training has not equipped these models with the crucial skill of actively seeking the user’s help when faced with problematic queries. This shortcoming is likely attributed to a lack of exposure to noisy and imperfect training data, which more accurately mirrors the complexities and ambiguities of real-world scenarios.

\paragraph{Training unlocks proactive critical thinking.}
% summarize
In a significant departure from vanilla performance, both SFT and RL yield substantial improvements in ACC and REQ. Notably, the Qwen3-1.7B and Llama-3.2-3B-Instruct models exhibit a remarkable 70\% increase on GSM-MC and a 40\% improvement on GSM-MCE after two-stage training. The performance gap with GSM8K is significantly narrowed.

Interestingly, the results for the Qwen3-8B model present an unexpected phenomenon: employing RL alone surpasses the performance of two-stage training. This may arise from the nature of the SFT data, which is self-generated by the Qwen3-8B model and thus does not inherently enhance its capabilities. Moreover, by further reinforcing its original high-probability tokens during SFT, the entropy of the model's outputs may be inadvertently reduced. This could constrain the exploratory nature of the subsequent RL phase, thereby hindering its overall effectiveness.

\paragraph{Training activates a beneficial ``thinking mode''.}
% \subfile{../tables/cases}
A notable observation from our experiments is that RL fundamentally changes how models use their internal ``thinking mode.'' For vanilla models, activating the ``thinking mode'' often degrades performance. The extended thinking appears to induce counterproductive self-doubt rather than useful analysis, leading to a clear drop in performance. For example, the Qwen3-8B model's ACC on GSM-MC decreases from 55.59\% to 44.92\% when enabling the ``thinking mode''.

However, this trend is decisively reversed after RL training. The ``thinking mode'' evolves into a powerful asset, enabling the model to carefully analyze flaws in prompts and formulate effective clarifying questions. This transformation is most evident in the Qwen3-8B, where enabling the ``thinking mode'' increases its ACC on GSM-MC from 81.73\% to a peak of 85.53\%.
Additionally, we observe that RL also enhances the model's thinking ability even without explicitly activating the ``thinking mode'', as evidenced by the increasing response length.
Representative cases further illustrating the profound impact of ``thinking mode'' are provided in Appendix~\ref{app:case}.

\input{tables/data_filter}

\paragraph{General capabilities are preserved.}
A critical outcome of our training is that the substantial gains in proactive critical thinking do not come at the cost of the models' core mathematical reasoning abilities. As presented in Table~\ref{tab:main_x}, the performance of our trained models on the standard GSM8K benchmark remains remarkably stable. In fact, for smaller models, training even yields slight improvements in their foundational skills. For instance, the Llama-3.2-3B-Instruct model's ACC on GSM8K increases significantly from 64.06\% to 75.74\% after undergoing the full SFT+RL training. This demonstrates that our approach successfully integrates a new, specialized skill of proactive critical thinking, while simultaneously preserving and even enhancing the models’ general capabilities, resulting in more robust and reliable models fit for complex, real-world applications.

\subsection{Analysis}
\subsubsection{The Importance of Data Filtering}

To verify the filtering process in our data curation method, we conduct an ablation study on Qwen3-1.7B model. In this experiment, we train a control version of the model using an unfiltered dataset created by randomly sampling from the original data pool. The size of this unfiltered dataset is kept identical to our curated, high-quality dataset to ensure a fair comparison. As detailed in Table~\ref{tab:data_filter}, removing the data filtering pipeline leads to a significant degradation in model performance. For instance, when the ``thinking mode'' enabled, the ACC of the SFT model on the GSM-MC benchmark falls sharply from 44.15\% to just 27.20\%. This substantial decrease underscores the criticality of our carefully designed data filtering process in enabling the model to acquire proactive critical thinking skills during training.

\begin{figure}[tp]
    \centering
    \includegraphics[width=\columnwidth]{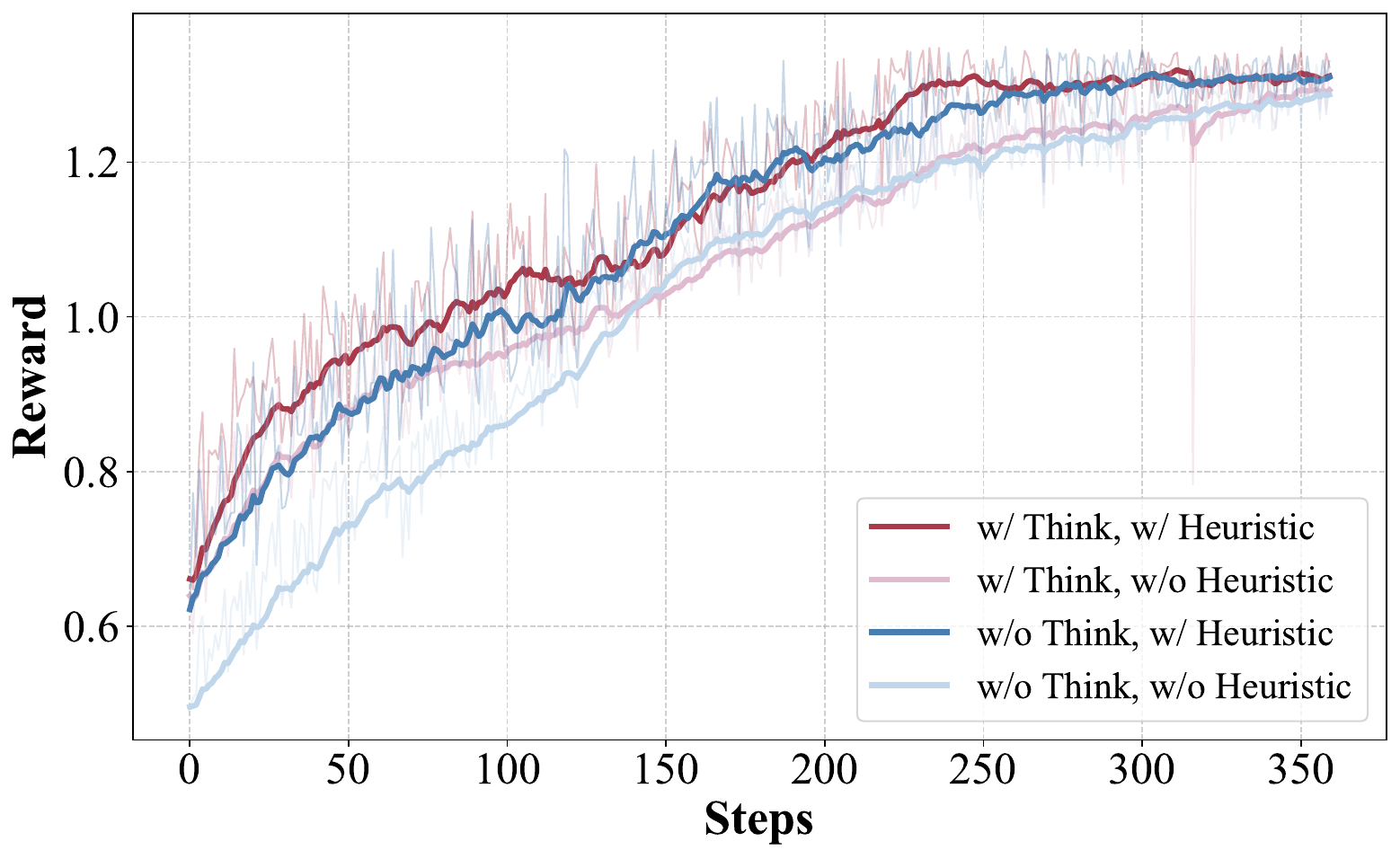}
    \caption{Reward curves in the RL stage following SFT, comparing the reward progression with and without incorporating heuristic information.}
    \label{fig:reward}
\end{figure}

\input{tables/ood}

\subsubsection{The Role of Heuristic Guidance}
To evaluate the influence of heuristic guidance during training, we analyze the reward curves in the RL phase. As shown in Figure~\ref{fig:reward}, models trained with heuristics (the red and blue lines) begin with a substantially higher initial reward compared to their counterparts trained without this guidance (the pink and light-blue lines). Furthermore, they exhibit a markedly faster rate of convergence, thereby achieving higher performance sooner. This distinct advantage persists regardless of whether the ``thinking mode'' is enabled, demonstrating that heuristic guidance serves as a powerful catalyst for more efficient RL.

\subsubsection{Generalization Analysis}
To assess the generalization capabilities of our trained models, we conduct an out-of-distribution evaluation using the MIP-MATH dataset~\cite{fan2025missing}. This challenging benchmark is derived from the standard MATH dataset~\cite{hendrycks2021measuring} by intentionally omitting a critical premise from each problem manually. The results, presented in Table~\ref{tab:ood}, demonstrate that our trained Qwen3-8B models significantly outperform the vanilla counterpart, irrespective of whether the ``thinking mode'' is enabled. This robust performance on the challenging, out-of-distribution benchmark indicates that the proactive critical thinking skills imparted by our method are not confined to the training distribution. Instead, our approach fosters a generalizable capability, further validating the effectiveness of our data curation and training paradigm.

\subsubsection{Multi-Turn Expansion Experiments}
\input{tables/multiturn}

Finally, we investigate whether the proactive critical thinking skills instilled in our models can generalize to a multi-turn dialogue setting, a scenario that is not explicitly included in training. To evaluate this, we design an experiment where the model must iteratively assess whether it has gathered sufficient information to answer, engaging in multiple rounds of clarification if necessary. The results, detailed in Table~\ref{tab:multiturn}, reveal a remarkable trend: model performance consistently improves as the number of interactive turns increases. For instance, the RL-trained Qwen3-8B model (with thinking enabled) improves its ACC from 85.53\% in a standard two-turn interaction to 88.67\% after four dialogue turns. This steady enhancement indicates that the learned proactive critical thinking is not a rigid, single-step capability but rather an adaptable skill that naturally extends to more complex interactions.

%% file: tables/main.tex
\begin{table}[t]
\small
\centering
\resizebox{\columnwidth}{!}{%
\begin{tabular}{@{}lcccccc@{}}
\begin{tabular}{l|cccccc} 
\toprule
\multicolumn{1}{c|}{\multirow{2}{*}{\textbf{Models}}} & \multicolumn{2}{c}{\textbf{GSM8K}}& \multicolumn{2}{c}{\textbf{GSM-MC}}& \multicolumn{2}{c}{\textbf{GSM-MCE}}\\ 
\cmidrule(l){2-3} \cmidrule(l){4-5} \cmidrule(l){6-7}
\multicolumn{1}{c|}{} & ACC             & REQ   & ACC           & REQ   & ACC           & REQ   \\ \midrule
\multicolumn{7}{c}{\textbf{Qwen3-1.7B w/o think}}\\ \midrule
Vanilla              & 78.01            & 0.00  & 0.95          & 1.75  & 0.62          & 1.15  \\
SFT                  & 78.32            & 6.90  & 38.60         & 77.70 & 17.46         & 69.40 \\
RL                   & 82.03            & 7.73  & 60.82         & 91.59 & 37.13         & 88.18 \\
SFT+RL               & \textbf{85.44}   & 6.75  & \textbf{73.68}& 95.69 & \textbf{40.04}& 90.12 \\ \midrule
\multicolumn{7}{c}{\textbf{Qwen3-1.7B w/ think}}\\ \midrule
Vanilla              & \textbf{87.79}   & 0.00  & 0.15          & 0.66  & 0.00          & 0.00  \\
SFT                  & 86.13            & 0.68  & 44.15         & 70.54 & 14.99         & 44.44 \\
RL                   & 87.41            & 5.46  & 62.13         & 92.54 & 29.19         & 84.22 \\
SFT+RL               & 85.75            & 5.91  & \textbf{73.98}& 97.00 & \textbf{41.09}& 92.95 \\ \midrule
\multicolumn{7}{c}{\textbf{Qwen3-8B w/o think}}\\ \midrule
Vanilla              & \textbf{92.19}   & 0.68  & 55.59         & 76.06 & 25.09         & 59.08 \\
SFT                  & 91.67            & 1.74  & 65.35         & 91.23 & 27.78         & 81.75 \\
RL                   & 90.60            & 5.53  & \textbf{81.73}& 99.63 & \textbf{42.59}& 95.86 \\
SFT+RL               & 91.21            & 4.55  & 79.24         & 99.12 & 36.16         & 95.33 \\ \midrule
\multicolumn{7}{c}{\textbf{Qwen3-8B w/ think}}\\ \midrule
Vanilla              & 94.62            & 0.08  & 44.92         & 57.75 & 14.11         & 29.59 \\
SFT                  & \textbf{95.60}   & 0.30  & 57.38         & 75.22 & 21.52         & 48.32 \\
RL                   & 92.27            & 5.00  & \textbf{85.53}& 99.12 & \textbf{49.38}& 94.80 \\
SFT+RL               & 93.48            & 3.26  & 83.11         & 98.61 & 41.62         & 92.86 \\ \midrule
\multicolumn{7}{c}{\textbf{Llama-3.2-3B-Instruct}}\\ \midrule
Vanilla              & 64.06            & 0.08  & 0.15          & 0.44  & 0.09          & 0.26  \\
SFT                  & 48.60            & 30.93 & 31.87         & 86.77 & 14.46         & 79.10 \\
RL                   & \textbf{79.15}   & 7.96  & 57.97         & 91.37 & 28.92         & 84.74 \\
SFT+RL               & 75.74            & 12.59 & \textbf{74.49}& 96.42 & \textbf{48.32}& 91.98 \\ 
\bottomrule
\end{tabular}
\end{tabular}%
}

\caption{Accuracy~(ACC) and Request Ratio~(REQ) of the models evaluated on the standard GSM8K and our two constructed benchmarks, GSM-MC and GSM-MCE.
For the Qwen3 models, we report results both with and without the ``thinking mode'' to examine the role of reasoning in this task.
% Results include the Qwen3 models with and without the ``thinking mode''.  
% For each dataset, the highest ACC is highlighted in \textbf{bold}.
}
\label{tab:main_x}
\end{table}

%% file: tables/data_filter.tex
\begin{table}[t]
\small
\centering
\resizebox{\columnwidth}{!}{%
\begin{tabular}{@{}lcccccc@{}}
\begin{tabular}{l|cccccc} 
\toprule
\multicolumn{1}{c|}{\multirow{2}{*}{\textbf{Models}}} & \multicolumn{2}{c}{\textbf{GSM8K}}& \multicolumn{2}{c}{\textbf{GSM-MC}}& \multicolumn{2}{c}{\textbf{GSM-MCE}}\\ 
\cmidrule(l){2-3} \cmidrule(l){4-5} \cmidrule(l){6-7}
\multicolumn{1}{c|}{} & ACC             & REQ   & ACC           & REQ   & ACC           & REQ   \\ \midrule
\multicolumn{7}{c}{\textbf{Qwen3-1.7B w/o think}}\\ \midrule
SFT                  & 78.32            & 6.90  & 38.60         & 77.70 & 17.46         & 69.40 \\
\ \ \textit{w/o filter}       & 76.65            & 5.23  & 34.71         & 68.93 &  14.98        & 57.93 \\
RL                   & 82.03            & 7.73  & 60.82         & 91.59 & 37.13         & 88.18 \\
\ \  \textit{w/o filter}      & 78.99            & 12.51 & 55.19         & 89.33 & 29.63         & 87.13 \\ \midrule
\multicolumn{7}{c}{\textbf{Qwen3-1.7B w/ think}}\\ \midrule
SFT                  & 86.13            & 0.68  & 44.15         & 70.54 & 14.99         & 44.44 \\
\ \ \textit{w/o filter}       & 90.75            & 0.38  & 27.20         & 48.25 & 9.44          & 27.34 \\
RL                   & 87.41            & 5.46  & 62.13         & 92.54 & 29.19         & 84.22 \\
\ \ \textit{w/o filter}       & 84.46            & 9.78  & 57.09         & 89.62 & 21.96         & 81.92 \\ 
\bottomrule
\end{tabular}
\end{tabular}%
}

\caption{Results of the ablation study on the data filtering process.}
\label{tab:data_filter}
\end{table}

%% file: tables/ood.tex
\begin{table}[t]
\small
\centering
\resizebox{\columnwidth}{!}{%
\begin{tabular}{l|cccc}
\toprule
\multirow{2}{*}{\textbf{Methods}} & 
\multicolumn{2}{c}{\textbf{Qwen3-8B w/o think}} & 
\multicolumn{2}{c}{\textbf{Qwen3-8B w/ think}} \\

 \cmidrule(l){2-3} \cmidrule(l){4-5} 
& ACC & REQ & ACC & REQ \\
\midrule
Vanilla     & 21.15 & 32.69 & 17.31 & 28.84 \\
SFT         & 42.31 & 73.08 & 26.92 & 46.15 \\
RL          & \textbf{57.69} & 80.77 & \textbf{59.62} & 76.92 \\
SFT+RL      & 50.00 & 80.77 & 55.77 & 78.85 \\
\bottomrule
\end{tabular}
}
\caption{Qwen3-8B model's performance on the out-of-distribution MIP-MATH dataset.}
\label{tab:ood}
\end{table}

%% file: tables/multiturn.tex
% \usepackage{booktabs}
% \usepackage{multirow}
% \usepackage{graphicx}
\begin{table}[t]
\small
\centering
% \resizebox{\columnwidth}{!}{%
\begin{tabular}{c|cccc}
\toprule
\multirow{2}{*}{\textbf{Turns}} & \multicolumn{2}{c}{\textbf{Qwen3-1.7B}} & \multicolumn{2}{c}{\textbf{Qwen3-8B}} \\ 
\cmidrule(l){2-3} \cmidrule(l){4-5}
   & w/o think & w/ think & w/o think & w/ think \\ 
\midrule

\multicolumn{5}{c}{\textbf{SFT}}\\ \midrule

 2 & 38.60             & 44.15            & 65.35             & 57.38 \\
 3 & 43.49             & 49.71            & 66.01             & 57.73 \\
 4 & \textbf{45.69}    & \textbf{50.64 }  & \textbf{66.52}    & \textbf{58.12} \\ 
\midrule
\multicolumn{5}{c}{\textbf{RL}}\\ \midrule

 2 & 60.82             & 62.13             & 81.73             & 85.53 \\
 3 & 61.26             & 62.72             & 83.26             & 88.30 \\
 4 & \textbf{61.99}    & \textbf{63.74}    & \textbf{84.06}    & \textbf{88.67} \\ 
\midrule

\multicolumn{5}{c}{\textbf{SFT+RL}}\\ \midrule

 2 & 73.68             & 73.98             & 79.24             & 83.11 \\
 3 & 74.63             & 74.49             & 80.48             & 85.96 \\
 4 & \textbf{74.93}    & \textbf{75.23}    & \textbf{81.07}    & \textbf{86.42} \\

\bottomrule
\end{tabular}%
% }
\caption{The ACC on the GSM-MC benchmark across multiple conversational turns.}
\label{tab:multiturn}
\end{table}

%% file: sections/6.Conclusion.tex
\section{Conclusion}

This work presents a preliminary study on proactive critical thinking in LLMs, where the model not only identifies flaws in user input but also actively engages with users to collaboratively solve problems.
We first propose a data preparation pipeline to construct GSM-MC and its more challenging extension, GSM-MCE, enabling systematic evaluation of proactive critical thinking. Building on these benchmarks, we then improve model performance through supervised fine-tuning and reinforcement learning, enhanced with heuristic guidance. Experimental results demonstrate that our approach leads to significant improvements on both test sets across three different models.
As this area is still underexplored, several promising directions remain:
(1) Developing benchmarks covering a broader range of domains, such as medicine.
(2) Extending interaction length to more turns to address more complex tasks.
(3) Exploring more robust training algorithms capable of providing diverse feedback beyond questioning.

%% file: tables/cases.tex
\begin{table*}[t]
\centering
\resizebox{\textwidth}{!}{%
\begin{tabular}{|m{0.50\textwidth}|m{0.50\textwidth}|}
\hline
\multicolumn{2}{|l|}{\parbox{\textwidth}{\textbf{Question:} Axel has 50 silver pesos. He visits his friend Anna who has twice as many silver pesos as he has and 40 more gold pesos. What's the total number of pesos they have together?}} \\
\hline
\textbf{(a) Vanilla Qwen3-8B w/o thinking} & \textbf{(b) Trained Qwen3-8B w/o thinking} \\
\hline
% 1-1
\textbf{Assistant:} First, let's break down the information ... Since the question asks for the total number of pesos they have together, but it does not specify how to handle the gold pesos (e.g., whether they are worth the same as silver pesos or not), we cannot compute the exact total without that information. Therefore, we need to ask: 
$$\boxed{\text{Are gold pesos equivalent in value to silver pesos?}}$$ \newline
\textbf{User:} No \newline
\textbf{Assistant:} Since the answer to the question ``Are gold pesos equivalent in value to silver pesos?'' is ``No'', it implies that gold pesos are not equivalent to silver pesos, or that their relationship is undefined... Axel has 50 silver pesos. Anna has 100 silver pesos. Anna also has 40 gold pesos, but since we don't know their value in silver, we cannot include them. So, the total number of silver pesos they have together is: 
$$50 + 100 = \boxed{150}$$
&
% 1-2
\textbf{Assistant:} To determine the total number of pesos Axel and Anna have together, we need to consider both silver pesos and gold pesos. \newline
Step 1: Determine how many silver pesos Anna has ... \newline
Step 2: ... Since we don't know how many gold pesos Axel has, we cannot calculate the total number of gold pesos they have together. Therefore, we need more information. 
$$\boxed{\text{What is the number of gold pesos Axel has?}}$$ \newline
\textbf{User:} 80 gold pesos \newline
\textbf{Assistant:} Axel has: 50 silver pesos, 80 gold pesos. Anna has:100 silver pesos (twice as many as Axel). 80 + 40 = 120 gold pesos (40 more than Axel).
\newline Total pesos together:\newline - Silver: $ 50 + 100 = 150 $\newline - Gold: $ 80 + 120 = 200 $  
$$ \text{Total} = 150 + 200 = \boxed{350} $$
\\
\hline
\textbf{(c) Vanilla Qwen3-8B w/ thinking} & \textbf{(d) Trained Qwen3-8B w/ thinking} \\
\hline
% 2-1
\textbf{Assistant:} \textless think\textgreater \ Okay, let's see. Axel has 50 silver pesos ... Wait, the question mentions ``gold pesos'' but the total is asked in terms of ``pesos'' ... Hmm, maybe the problem is considering that gold pesos are also counted as pesos? ... So maybe the question is considering that gold pesos are a different type, but the total is just the sum of silver and gold? ... Hmm. Wait, maybe the question is in a context where all pesos, whether silver or gold, are just considered as pesos ... Alternatively, maybe the question has a typo, and ``gold pesos'' is supposed to be ``silver pesos'' ... So, Axel has 50 silver, Anna has 100 silver and 40 gold. So total is $50 + 100 + 40 = 190$. So, the answer would be $50 + 100 + 40 = 190$. Let me check again ... I think the answer is 190. So, I'll go with that. \textless /think\textgreater \newline
To determine the total number of pesos that Axel and Anna have together, we need to account for all the pesos each of them possesses. Axel has 50 silver pesos. Anna has twice as many silver pesos as Axel, which is $ 2 \times 50 = 100 $ silver pesos. Additionally, Anna has 40 more gold pesos. The question asks for the total number of pesos they have together. The total number of pesos is:
$$50 + 100 + 40 = \boxed{190} $$
&

% 2-2
\textbf{Assistant:} \textless think\textgreater \ Okay, let's see. Axel has 50 silver pesos ... Wait, but Axel's gold isn't mentioned ... But how much does Axel have in gold? The problem doesn't mention Axel's gold. So, maybe the question is missing some information? ... Therefore, we can't solve it as it is. So, we need to ask the user for the amount of gold pesos that Axel has ... So, the final answer is to ask for the necessary information. \textless /think\textgreater \newline
The problem states that ... However, it does not specify how many gold pesos Axel has, nor does it clarify whether gold pesos should be included in the total number of pesos ... It is impossible to compute the exact total. $$\boxed{\text{How much does Axel have in gold pesos?}}$$ \newline
\textbf{User:} Axel has 80 gold pesos. \newline
\textbf{Assistant:} Given that Axel has 80 gold pesos, we can now compute the total number of pesos that Axel and Anna have together: Axel's silver pesos: 50. Axel's gold pesos: 80. Anna's silver pesos: Twice as many as Axel's silver, $2 \times 50 = 100$. Anna's gold pesos: 40 more than Axel's gold = $80 + 40 = 120$. Now, summing up all the pesos:  
$$50 + 80 + 100 + 120 =\boxed{350}$$ \\
\hline
\end{tabular}
}
\caption{Representative cases comparing the vanilla Qwen3-8B model with its trained counterpart.}
\label{tab:cases}
\end{table*}

%% file: Formatting-Instructions-LaTeX-2026.bbl
\begin{thebibliography}{32}
\providecommand{\natexlab}[1]{#1}

\bibitem[{Alkaabi and Elsori(2025)}]{alkaabi2025navigating}
Azza Alkaabi and Deena Elsori. 2025.
\newblock Navigating digital frontiers in uae healthcare: A qualitative exploration of healthcare professionals’ and patients’ experiences with ai and telemedicine.
\newblock \emph{PLOS Digital Health}, 4(4):e0000586.

\bibitem[{Andukuri et~al.(2024)Andukuri, Fr{\"a}nken, Gerstenberg, and Goodman}]{andukuri2024star}
Chinmaya Andukuri, Jan-Philipp Fr{\"a}nken, Tobias Gerstenberg, and Noah~D Goodman. 2024.
\newblock Star-gate: Teaching language models to ask clarifying questions.
\newblock \emph{arXiv preprint arXiv:2403.19154}.

\bibitem[{Cobbe et~al.(2021)Cobbe, Kosaraju, Bavarian, Chen, Jun, Kaiser, Plappert, Tworek, Hilton, Nakano, Hesse, and Schulman}]{cobbe2021gsm8k}
Karl Cobbe, Vineet Kosaraju, Mohammad Bavarian, Mark Chen, Heewoo Jun, Lukasz Kaiser, Matthias Plappert, Jerry Tworek, Jacob Hilton, Reiichiro Nakano, Christopher Hesse, and John Schulman. 2021.
\newblock Training verifiers to solve math word problems.
\newblock \emph{arXiv preprint arXiv:2110.14168}.

\bibitem[{Dong et~al.(2024)Dong, Jiang, Jin, and Li}]{dong2024self}
Yihong Dong, Xue Jiang, Zhi Jin, and Ge~Li. 2024.
\newblock Self-collaboration code generation via chatgpt.
\newblock \emph{ACM Transactions on Software Engineering and Methodology}, 33(7):1--38.

\bibitem[{Fan et~al.(2025)Fan, Li, Sun, and Zhou}]{fan2025missing}
Chenrui Fan, Ming Li, Lichao Sun, and Tianyi Zhou. 2025.
\newblock Missing premise exacerbates overthinking: Are reasoning models losing critical thinking skill?
\newblock \emph{arXiv preprint arXiv:2504.06514}.

\bibitem[{Feng et~al.(2023)Feng, Vanam, Cherukupally, Zheng, Qiu, and Chen}]{feng2023investigating}
Yunhe Feng, Sreecharan Vanam, Manasa Cherukupally, Weijian Zheng, Meikang Qiu, and Haihua Chen. 2023.
\newblock Investigating code generation performance of chatgpt with crowdsourcing social data.
\newblock In \emph{2023 IEEE 47th Annual Computers, Software, and Applications Conference (COMPSAC)}, pages 876--885. IEEE.

\bibitem[{Hendrycks et~al.(2021)Hendrycks, Burns, Kadavath, Arora, Basart, Tang, Song, and Steinhardt}]{hendrycks2021measuring}
Dan Hendrycks, Collin Burns, Saurav Kadavath, Akul Arora, Steven Basart, Eric Tang, Dawn Song, and Jacob Steinhardt. 2021.
\newblock \href {https://datasets-benchmarks-proceedings.neurips.cc/paper_files/paper/2021/file/be83ab3ecd0db773eb2dc1b0a17836a1-Paper-round2.pdf} {Measuring mathematical problem solving with the math dataset}.
\newblock In \emph{Proceedings of the Neural Information Processing Systems Track on Datasets and Benchmarks}, volume~1.

\bibitem[{IMANI et~al.(2024)IMANI, Shrivastava, and Du}]{imani2024mathematical}
Shima IMANI, Harsh Shrivastava, and Liang Du. 2024.
\newblock Mathematical reasoning using large language models.
\newblock US Patent App. 18/144,802.

\bibitem[{Kim et~al.(2024)Kim, Lee, Kim, Park, and Kim}]{kim2024understanding}
Yoonsu Kim, Jueon Lee, Seoyoung Kim, Jaehyuk Park, and Juho Kim. 2024.
\newblock Understanding users’ dissatisfaction with chatgpt responses: Types, resolving tactics, and the effect of knowledge level.
\newblock In \emph{Proceedings of the 29th International Conference on Intelligent User Interfaces}, pages 385--404.

\bibitem[{Kirichenko et~al.(2025)Kirichenko, Ibrahim, Chaudhuri, and Bell}]{kirichenko2025abstentionbench}
Polina Kirichenko, Mark Ibrahim, Kamalika Chaudhuri, and Samuel~J Bell. 2025.
\newblock Abstentionbench: Reasoning llms fail on unanswerable questions.
\newblock \emph{arXiv preprint arXiv:2506.09038}.

\bibitem[{Kuhn et~al.(2022)Kuhn, Gal, and Farquhar}]{kuhn2022clam}
Lorenz Kuhn, Yarin Gal, and Sebastian Farquhar. 2022.
\newblock Clam: Selective clarification for ambiguous questions with generative language models.
\newblock \emph{arXiv preprint arXiv:2212.07769}.

\bibitem[{Li et~al.(2024)Li, Cui, Zhao, Kong, and Bi}]{li2024gsm}
Qintong Li, Leyang Cui, Xueliang Zhao, Lingpeng Kong, and Wei Bi. 2024.
\newblock Gsm-plus: A comprehensive benchmark for evaluating the robustness of llms as mathematical problem solvers.
\newblock In \emph{Proceedings of the 62nd Annual Meeting of the Association for Computational Linguistics (Volume 1: Long Papers)}, pages 2961--2984.

\bibitem[{Li et~al.(2025)Li, Mun, Brahman, Ilgen, Tsvetkov, and Sap}]{li2025aligning}
Shuyue~Stella Li, Jimin Mun, Faeze Brahman, Jonathan Ilgen, Yulia Tsvetkov, and Maarten Sap. 2025.
\newblock \href {https://doi.org/10.48550/arXiv.2502.14860} {Aligning llms to ask good questions a case study in clinical reasoning}.
\newblock \emph{CoRR}, abs/2502.14860.

\bibitem[{Lin et~al.(2025)Lin, Wang, Chen, Liu, Liu, Su, and Xiao}]{lin-etal-2025-investigating-inference}
Yujie Lin, Ante Wang, Moye Chen, Jingyao Liu, Hao Liu, Jinsong Su, and Xinyan Xiao. 2025.
\newblock \href {https://aclanthology.org/2025.findings-acl.808/} {Investigating inference-time scaling for chain of multi-modal thought: A preliminary study}.
\newblock In \emph{Findings of the Association for Computational Linguistics: ACL 2025}, pages 15654--15667, Vienna, Austria. Association for Computational Linguistics.

\bibitem[{Liu et~al.(2024)Liu, Feng, Xue, Wang, Wu, Lu, Zhao, Deng, Zhang, Ruan et~al.}]{liu2024deepseek}
Aixin Liu, Bei Feng, Bing Xue, Bingxuan Wang, Bochao Wu, Chengda Lu, Chenggang Zhao, Chengqi Deng, Chenyu Zhang, Chong Ruan, et~al. 2024.
\newblock Deepseek-v3 technical report.
\newblock \emph{arXiv preprint arXiv:2412.19437}.

\bibitem[{Ma et~al.(2024)Ma, Dai, Yuan, Luo, Wang, Liu, Sha, Sui et~al.}]{ma2024large}
Jingyuan Ma, Damai Dai, Zihang Yuan, Weilin Luo, Bin Wang, Qun Liu, Lei Sha, Zhifang Sui, et~al. 2024.
\newblock Large language models struggle with unreasonability in math problems.
\newblock \emph{arXiv preprint arXiv:2403.19346}.

\bibitem[{Meta(2024)}]{meta2024llama}
AI~Meta. 2024.
\newblock Llama 3.2: Revolutionizing edge ai and vision with open, customizable models.
\newblock \emph{Meta AI Blog. Retrieved December}, 20:2024.

\bibitem[{Rahman et~al.(2024)Rahman, Ye, Yao, Liu, Yu, Yu, Yin, and Wang}]{rahman2024blind}
AM~Rahman, Junyi Ye, Wei Yao, Sierra~S Liu, Jesse Yu, Jonathan Yu, Wenpeng Yin, and Guiling Wang. 2024.
\newblock From blind solvers to logical thinkers: Benchmarking llms' logical interity on faulty mathematical problems.
\newblock \emph{arXiv preprint arXiv:2410.18921}.

\bibitem[{Shao et~al.(2025)Shao, Yan, Poshyvanyk, and Su}]{shao2025code}
Liangying Shao, Yanfu Yan, Denys Poshyvanyk, and Jinsong Su. 2025.
\newblock \href {https://doi.org/10.1109/ICSE-NIER66352.2025.00020} {Unigencoder: Merging seq2seq and seq2tree paradigms for unified code generation}.
\newblock In \emph{2025 IEEE/ACM 47th International Conference on Software Engineering: New Ideas and Emerging Results (ICSE-NIER)}, pages 71--75.

\bibitem[{Shao et~al.(2024)Shao, Wang, Zhu, Xu, Song, Bi, Zhang, Zhang, Li, Wu et~al.}]{shao2024deepseekmath}
Zhihong Shao, Peiyi Wang, Qihao Zhu, Runxin Xu, Junxiao Song, Xiao Bi, Haowei Zhang, Mingchuan Zhang, YK~Li, Yang Wu, et~al. 2024.
\newblock Deepseekmath: Pushing the limits of mathematical reasoning in open language models.
\newblock \emph{arXiv preprint arXiv:2402.03300}.

\bibitem[{Song et~al.(2025)Song, Shi, and Zhao}]{song2025hallucination}
Linxin Song, Taiwei Shi, and Jieyu Zhao. 2025.
\newblock The hallucination tax of reinforcement finetuning.
\newblock \emph{arXiv preprint arXiv:2505.13988}.

\bibitem[{Sun et~al.(2024)Sun, Yin, Guo, Wu, Qiu, and Zhao}]{sun2024benchmarking}
YuHong Sun, Zhangyue Yin, Qipeng Guo, Jiawen Wu, Xipeng Qiu, and Hui Zhao. 2024.
\newblock Benchmarking hallucination in large language models based on unanswerable math word problem.
\newblock In \emph{Proceedings of the 2024 Joint International Conference on Computational Linguistics, Language Resources and Evaluation (LREC-COLING 2024)}, pages 2178--2188.

\bibitem[{Wang et~al.(2025{\natexlab{a}})Wang, Song, Tian, Peng, Yu, Mi, Su, and Yu}]{wang2025litesearch}
Ante Wang, Linfeng Song, Ye~Tian, Baolin Peng, Dian Yu, Haitao Mi, Jinsong Su, and Dong Yu. 2025{\natexlab{a}}.
\newblock Litesearch: Efficient tree search with dynamic exploration budget for math reasoning.
\newblock In \emph{Proceedings of the AAAI Conference on Artificial Intelligence}, volume~39, pages 25318--25326.

\bibitem[{Wang et~al.(2024)Wang, Shi, Ling, Chan, Wang, Lee, Yuan, Huang, Jiao, and Lyu}]{wang2024learning}
Wenxuan Wang, Juluan Shi, Zixuan Ling, Yuk-Kit Chan, Chaozheng Wang, Cheryl Lee, Youliang Yuan, Jen-tse Huang, Wenxiang Jiao, and Michael~R Lyu. 2024.
\newblock Learning to ask: When llm agents meet unclear instruction.
\newblock \emph{arXiv preprint arXiv:2409.00557}.

\bibitem[{Wang et~al.(2025{\natexlab{b}})Wang, Li, Zhang, Wu, Liu, Hu, Guo, Huang, Xin, Yang, Su, Chen, and Li}]{wang2025epicoder}
Yaoxiang Wang, Haoling Li, Xin Zhang, Jie Wu, Xiao Liu, Wenxiang Hu, Zhongxin Guo, Yangyu Huang, Ying Xin, Yujiu Yang, Jinsong Su, Qi~Chen, and Scarlett Li. 2025{\natexlab{b}}.
\newblock \href {https://openreview.net/forum?id=RAxe7nF4Oz} {Epicoder: Encompassing diversity and complexity in code generation}.
\newblock In \emph{Forty-second International Conference on Machine Learning}.

\bibitem[{Wu et~al.(2025)Wu, Galley, Peng, Cheng, Li, Dou, Cai, Zou, Leskovec, and Gao}]{wu2025collabllm}
Shirley Wu, Michel Galley, Baolin Peng, Hao Cheng, Gavin Li, Yao Dou, Weixin Cai, James Zou, Jure Leskovec, and Jianfeng Gao. 2025.
\newblock \href {https://openreview.net/forum?id=DmH4HHVb3y} {Collab{LLM}: From passive responders to active collaborators}.
\newblock In \emph{Forty-second International Conference on Machine Learning}.

\bibitem[{Yamauchi et~al.(2023)Yamauchi, Sonoda, Sannai, and Kumagai}]{yamauchi2023lpml}
Ryutaro Yamauchi, Sho Sonoda, Akiyoshi Sannai, and Wataru Kumagai. 2023.
\newblock Lpml: llm-prompting markup language for mathematical reasoning.
\newblock \emph{arXiv preprint arXiv:2309.13078}.

\bibitem[{Yang et~al.(2025)Yang, Li, Yang, Zhang, Hui, Zheng, Yu, Gao, Huang, Lv et~al.}]{yang2025qwen3}
An~Yang, Anfeng Li, Baosong Yang, Beichen Zhang, Binyuan Hui, Bo~Zheng, Bowen Yu, Chang Gao, Chengen Huang, Chenxu Lv, et~al. 2025.
\newblock Qwen3 technical report.
\newblock \emph{arXiv preprint arXiv:2505.09388}.

\bibitem[{Yao et~al.(2023)Yao, Yu, Zhao, Shafran, Griffiths, Cao, and Narasimhan}]{yao2023tree}
Shunyu Yao, Dian Yu, Jeffrey Zhao, Izhak Shafran, Tom Griffiths, Yuan Cao, and Karthik Narasimhan. 2023.
\newblock Tree of thoughts: Deliberate problem solving with large language models.
\newblock \emph{Advances in neural information processing systems}, 36:11809--11822.

\bibitem[{Zamfirescu-Pereira et~al.(2023)Zamfirescu-Pereira, Wong, Hartmann, and Yang}]{zamfirescu2023johnny}
J~Diego Zamfirescu-Pereira, Richmond~Y Wong, Bjoern Hartmann, and Qian Yang. 2023.
\newblock Why johnny can’t prompt: how non-ai experts try (and fail) to design llm prompts.
\newblock In \emph{Proceedings of the 2023 CHI conference on human factors in computing systems}, pages 1--21.

\bibitem[{Zhang et~al.(2025)Zhang, Knox, and Choi}]{zhang2024modeling}
Michael~JQ Zhang, W.~Bradley Knox, and Eunsol Choi. 2025.
\newblock \href {https://openreview.net/forum?id=cwuSAR7EKd} {Modeling future conversation turns to teach {LLM}s to ask clarifying questions}.
\newblock In \emph{The Thirteenth International Conference on Learning Representations}.

\bibitem[{Zhou et~al.(2024)Zhou, Song, Yao, Shu, and Ma}]{zhou2024isr}
Zhehua Zhou, Jiayang Song, Kunpeng Yao, Zhan Shu, and Lei Ma. 2024.
\newblock Isr-llm: Iterative self-refined large language model for long-horizon sequential task planning.
\newblock In \emph{2024 IEEE International Conference on Robotics and Automation (ICRA)}, pages 2081--2088. IEEE.

\end{thebibliography}
